\documentclass[sigconf]{acmart}
\usepackage{subcaption}
\AtBeginDocument{%
  \providecommand\BibTeX{{%
    \normalfont B\kern-0.5em{\scshape i\kern-0.25em b}\kern-0.8em\TeX}}}

\copyrightyear{2022} 
\acmYear{2022} 
\setcopyright{acmlicensed}\acmConference[GECCO '22]{Genetic and Evolutionary Computation Conference}{July 9--13, 2022}{Boston, MA, USA}
\acmBooktitle{Genetic and Evolutionary Computation Conference (GECCO '22), July 9--13, 2022, Boston, MA, USA}
\acmPrice{15.00}
\acmDOI{10.1145/3512290.3528751}
\acmISBN{978-1-4503-9237-2/22/07}

\newcommand{\hbr}{HTE}

\begin{document}

\title{Hierarchical Quality-Diversity for Online Damage Recovery}


\author{Maxime Allard}
\affiliation{%
   \department{Adaptive and Intelligent Robotics Lab}
  \institution{Imperial College London \country{United Kingdom}}
}
\email{maxime.allard@imperial.ac.uk}

\author{Simón C. Smith}
\affiliation{%
   \department{Adaptive and Intelligent Robotics Lab}
  \institution{Imperial College London \country{United Kingdom}}
  }
\email{s.smith-bize@imperial.ac.uk}

\author{Konstantinos Chatzilygeroudis}
\affiliation{%
\department{Computer Engineering and Informatics Department}
  \institution{University of Patras \country{Greece}}
  }
\email{costashatz@upatras.gr}

\author{Antoine Cully}
\affiliation{%
  \department{Adaptive and Intelligent Robotics Lab}
  \institution{Imperial College London \country{United Kingdom}}
  }
\email{a.cully@imperial.ac.uk}

\begin{abstract}
Adaptation capabilities, like damage recovery, are crucial for the deployment of robots in complex environments. Several works have demonstrated that using repertoires of pre-trained skills can enable robots to adapt to unforeseen mechanical damages in a few minutes. 
These adaptation capabilities are directly linked to the behavioural diversity in the repertoire. The more alternatives the robot has to execute a skill, the better are the chances that it can adapt to a new situation. However, solving complex tasks, like maze navigation, usually requires multiple different skills. Finding a large behavioural diversity for these multiple skills often leads to an intractable exponential growth of the number of required solutions. 
In this paper, we introduce the Hierarchical Trial and Error algorithm, which uses a hierarchical behavioural repertoire to learn diverse skills and leverages them to make the robot more adaptive to different situations. We show that the hierarchical decomposition of skills enables the robot to learn more complex behaviours while keeping the learning of the repertoire tractable. The experiments with a hexapod robot show that our method solves maze navigation tasks with 20\% less actions in the most challenging scenarios than the best baseline while having 57\% less complete failures.
\end{abstract}

\ccsdesc[500]{Computing methodologies~Evolutionary robotics}

\keywords{Hierarchical Learning, Quality-Diversity, Robotics}


\begin{teaserfigure}
\centering
\includegraphics[width=1.0\textwidth]{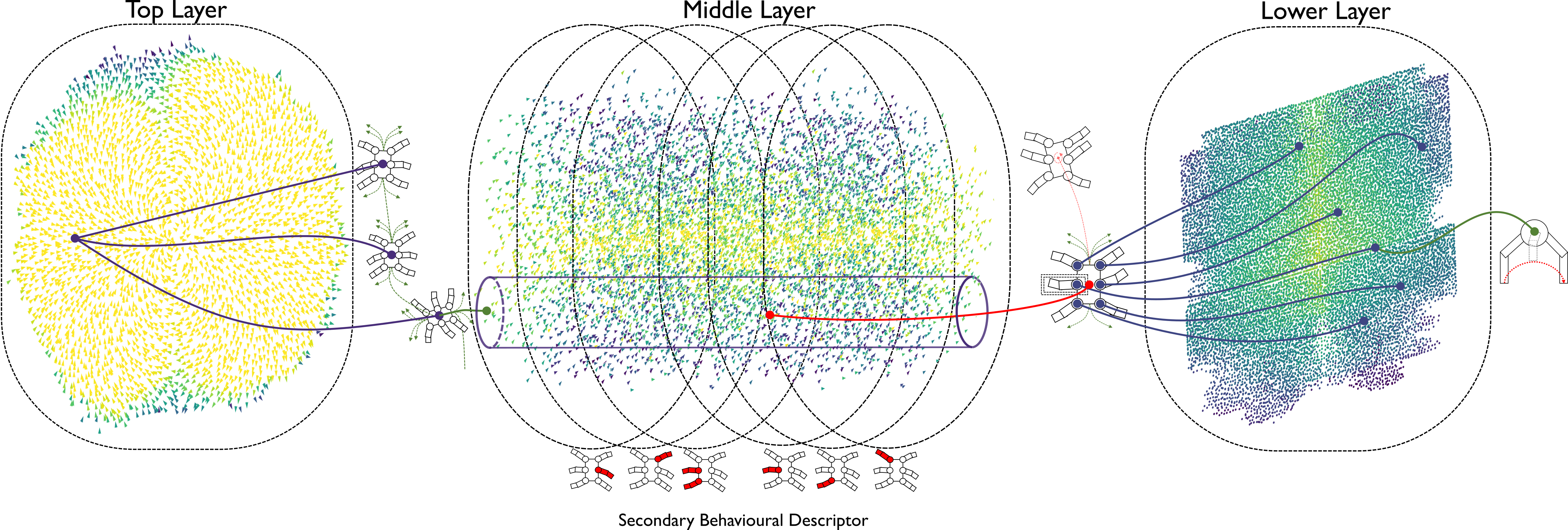}
\centering
\caption{Schematic representation of a 3-layered Hierarchical Trial and Error repertoire for omni-directional hexapod locomotion. The top behavioural repertoire (BR) fetches three different solutions from the middle level BR which in turn fetches each leg controller for the robot individually to achieve the desired upper-level behaviour. The middle layer has a depth (represented by a cylinder) from which the secondary behaviour can be chosen.}
\label{fig:hbr_architecture}
\end{teaserfigure}

\maketitle
\section{Introduction}

Robot learning has significantly advanced in the last years~\cite{miki2022learning,akkaya2019solving,Cully} by combining powerful optimisation methods, including Deep Learning, Reinforcement Learning and Quality-Diversity (QD)~\cite{shrestha2019review,sutton2018reinforcement,Long2016,Chatzilygeroudis2020Quality-DiversityOptimization}. However, the deployment of robots in realistic scenarios is still a hard problem~\cite{chatzilygeroudis2019survey}. Some of the problems that robots need to tackle before deployment include robustness to external perturbations, generalisation of  experience in novel scenarios and interaction with other complex agents (both artificial and natural), among several others~\cite{thrun2002probabilistic}. 
As it is impossible for a robot to be trained for all possible situations, it is possible to either generate relevant scenarios automatically beforehand \cite{Fontaine2020,Gambi2019,Rocklage2017,Mullins2017} or to adapt to them during deployment. Instead of relying on a single fixed control policy, a robot that adapts has the potential to perform efficiently in previously unseen scenarios~\cite{aastrom2013adaptive}.
Adaptive control has been proposed as a way of tuning parameters during exploitation~\cite{aastrom2013adaptive,cameron1984self}. In autonomous robots, the parameters of the controller can be directly updated based on the error of a predictive model to maintain an exploratory behaviour~\cite{smith2020diamond, 9618829}. Another way to achieve adaptation is to leverage multiple diverse control policies. For example, methods that leverage Quality-Diversity algorithms have shown that diversity of pre-computed solutions is a key factor for fast adaptation~\cite{Cully,Chatzilygeroudis2018Reset-freeRecovery,Kaushik2020AdaptiveRobotics}.

Quality-Diversity (QD) \cite{Long2016,7959075,Chatzilygeroudis2020Quality-DiversityOptimization} algorithms are a powerful family of algorithms that can be used to find a repertoire with a large set of diverse and high-performing robotic skills. 
In the work of Cully et al., the authors present the Intelligent Trial and Error algorithm (ITE) ~\cite{Cully} which uses QD to train around 13,000 solutions for a hexapod robot to walk in different directions. In a scenario where the robot has a damaged leg, ITE uses its large repertoire with a diverse set of controllers to find the solutions that are unaffected by the leg damage. 
The algorithm updates a set of Gaussian processes with experiences gathered by trial and error in the environment to find the best solution in the pre-trained repertoire. 
Other approaches to repertoire-based adaptation with QD include the Reset-Free Trial and Error algorithm (RTE) ~\cite{Chatzilygeroudis2018Reset-freeRecovery} and Adaptive Prior Selection for Repertoire-Based Online Learning algorithm (APROL) \cite{Kaushik2020AdaptiveRobotics}.
To solve navigation tasks with a damaged robot, RTE and APROL generate diverse controllers that are used for planning. APROL takes advantage of multiple trained repertoires with different prior knowledge on possible future conditions, e.g. damage or change in the friction between the robot and the ground, whereas RTE only uses a single repertoire.

All of the approaches above assume that a subset of the pre-trained solutions in a repertoire can be used to achieve adaptation.
The stochasticity of the environment and the complexity of the tasks may break this assumption if the diversity of solutions is not large enough.
One approach to tackle this problem is to increase the number of solutions in the repertoire, but at the cost of sampling efficiency, memory storage and a larger solution space. 

To increase the diversity of solutions for robot control, it is possible to decompose the search-space as a hierarchy and efficiently organise controllers~\cite{Merel2019}. The different layers allow for compositionality and re-usability of solutions (i.e. controllers). Furthermore, the use of hierarchies can help with transferability to other domains and few-shot learning~\cite{eppe2020hierarchical,Cully2018,Etcheverry}. In this work, we propose to leverage a hierarchical structure of repertoires with controllers to increase the diversity of solutions and improve the overall adaptability of a robot.

To this end, we introduce the Hierarchical Trial and Error (HTE) algorithm ,which uses hierarchical behavioural repertoires (HBR) \cite{Cully2018}. HBRs are composed of layers with different repertoires of solutions. Each repertoire stores solutions that range from low-level motor commands to high-level task goals descriptions. In HBRs, the layers are connected by one or more edges, implemented as the genotype of each solution.
The different layers of \hbr{} are trained to exhibit different behaviours on the robot (e.g. move a leg, walk for 1 second, etc). By including new behavioural descriptor dimensions, i.e. a secondary behavioural descriptors in the layers of \hbr{}, it is possible to increase the diversity of the solutions during the deployment.
Following RTE, our method uses the experience generated by the robot during the deployment phase to update a set of Gaussian processes. The Gaussian processes are used in a planning step with an MCTS algorithm to find the best action in the hierarchical repertoire based on the current state of the robot and the goal of the task. 

To test our hierarchical algorithm, we compare it to  RTE and APROL in a set of experiments with a hexapod robot in a simulated maze environment. We want to answer the following questions around the usefulness of \hbr{}:
\begin{itemize}
\item Is it possible to find diverse and useful controllers for omni-directional hexapod locomotion with \hbr{}?
\item Are we able to gain a diversity of skills by using an additional secondary behavioural descriptors in the layers?
\item Can \hbr{} solve complex downstream robot tasks? and, how does it compare to other baselines such as RTE and APROL?
\end{itemize}

Our results show that \hbr{} can find diverse solutions by decomposing the search space hierarchically. This makes the optimisation process simpler in comparison to finding solutions in the full extended solution space.
In comparison to the baselines, \hbr{} is able to be more adaptive (e.g. require less steps to solve the task) and less failures in a maze navigation task (i.e. downstream task) since it can effectively leverage the large collection of hierarchical skills.

\section{Background and Related work}

\textit{\textbf{Quality-Diversity.}}
Quality-Diversity has shown to be a powerful optimisation tool that can solve complex problems such as robot damage recovery fast and reliably~\cite{Cully} , achieving state-of-the-art results on unsolved Reinforcement Learning tasks in sparse-reward environments \cite{Ecoffet2021FirstExplore} , human-robot interaction~\cite{fontaine2020quality}, robotics~\cite{nordmoen2021map,Grillotti2021UnsupervisedOptimisation,eysenbach2018diversity,Mouret2020QualityOptimization}, games~\cite{perez2021generating,sarkar2021generating,steckel2021illuminating} and optimisation~\cite{fioravanzomap,salehi2021br} among several others.

Quality-Diversity (QD) algorithms ~\cite{Long2016,Chatzilygeroudis2020Quality-DiversityOptimization} are a family of evolutionary algorithms that aim to generate a collection of diverse and high-performing solutions. The diversity of solutions is obtained by defining a \emph{behavioural descriptor} $\mathbf{bd}$ (also known as feature vector)  used to characterise a solution. This diversity of solutions contrasts with classical optimisation algorithms that only look for the highest performing solution. For example, a mobile agent can reach a final destination by following different (diverse) paths. Each one of these paths is a valid solution to reach the target but they are all different in how they reach it.
In QD algorithms, the expert defines a $\emph{behavioural\ space}\ B \in \mathbb{R}^n$ with $n$-dimensions to describe a solution. After evaluation, each solution $\theta$ is associated with a behavioural descriptor $\mathbf{bd_\theta}$ that defines their behaviour in the space $B$.
Usually, the engineer defines the behavioural descriptor $\mathbf{bd_\theta}$ manually, which requires expert knowledge and can result in a solution bias during the evolutionary process. To define the behavioural descriptor automatically, recent methods encode the behaviours into a latent space with dimensionality-reduction algorithms ~\cite{Cully2018,Grillotti2021UnsupervisedOptimisation,Cully2019,PaoloUnsupervisedSpace,Laversanne-FinotCuriositySpaces} to discover behaviours in the latent space.

The solutions $\theta$ are defined by a genotype $\mathbf{g}$, which belongs to the $\emph{genotype\ space}\ G \in \mathbb{R}^k$ with $k$-dimensions. 
The most popular QD algorithms include MAP-Elites~\cite{Mouret2015} and Novelty Search~\cite{Lehman2011}. We use MAP-Elites in our method to generate the repertoires.
In MAP-Elites, once a solution has been evaluated, it is stored in a grid-like archive, called repertoire. The repertoire discretises the $\emph{behavioural\ space}\ B$ to store individuals in distinct cells.
In the case that two solutions have the same behavioural descriptor, the algorithm keeps the one with the highest fitness in the repertoire and discards the other.

\textit{\textbf{Hierarchical Organisation of Autonomous Control.}}
In nature, the nervous system of complex organisms, e.g. mammals, use hierarchical controllers for robust and versatile behaviours \cite{Merel2019}. Even more, authors in cognitive psychology have deemed hierarchies as critical mechanisms for developing intelligent agents~\cite{eppe2020hierarchical}.

Hierarchical structures allow algorithms to compose complex solutions out of primitive ones, which helps with the optimisation process.
For example, hierarchies are used to stack different levels of abstracted goals to solve high-dimensional problems in reinforcement learning ~\cite{Gehring2021HierarchicalExploration,Nachum2018} or to enable hierarchical latent spaces to discover a diversity of behavioural representations in evolutionary algorithms~\cite{Etcheverry}.

Furthermore, Hierarchical Genetic Algorithms \cite{Ryan2020Pyramid:Algorithms} use hierarchies with fitness functions of different granularities per layer, allowing the evolution of both coarse and fine behaviours at different levels.
Also, hierarchical abstractions are effective at controlling robots in more complex environments.
Several works show that the hierarchical abstraction of behaviours allows robots to solve complex tasks ~\cite{Duarte2016,Jain2020FromLocomotion,Li2021PlanningLocomotion}. In these works, a neural network uses planning to choose the actions from a repertoire of  primitive behaviours. 

Quality-Diversity has been successfully combined with hierarchical structures for a rapid divergent search of solutions with trees \cite{Smith2016} or to create hierarchical behavioural repertoires (HBR) that can be transferred across different types of robots to solve complex tasks such as drawing digits with a robotic arm \cite{Cully2018}. Our method is inspired by this latter work on HBRs and reuses the concept to create a diverse set of solutions. 



\textit{\textbf{QD-Based Adaptation.}}
Adaptability is a key feature for controlling robots in realistic scenarios. Usually, real-world scenarios define intractable state and solution spaces. It is impossible for a robot to have a pre-defined solution for each possible situation. Thus, the robot has to adapt to any occurring situation in real-time. 

Cully et al. introduced the Intelligent Trial and Error (ITE) algorithm as an adaptive control method based on MAP-Elites and Gaussian processes for planning and adaptation~\cite{Cully}. The most significant results of ITE include the rapid adaptation of a hexapod robot with damaged legs. Some other works build upon ITE to introduce a local adaptation mechanism to improve the simulation-to-reality transfer ~\cite{kim2019exploration}, to optimise a swarm of robots ~\cite{bossens2020rapidly} or to find new game levels with different difficulties \cite{Duque2020}. 
Along the same lines, the Adaptive Prior Selection for Repertoire-Based Online Learning algorithm (APROL)~\cite{Kaushik2020AdaptiveRobotics}, adapts the behaviour of a hexapod to different damages while executing a navigation task. To do so, the algorithm chooses the best skill, using Gaussian processes, among a set of dozens of repertoires of solutions that have been trained on potential scenarios (e.g. damage to the legs or different friction coefficients) that the robot could face during deployment. In an iterative process, APROL selects the best solutions from the repertoire that is the most likely to represent the actual conditions (i.e. friction coefficient or leg damage). 
These methods show that the solutions created by QD are useful to adapt to different situations while executing a task. Similarly, the Hierarchical Trial and Error algorithm is built upon the idea to create a diverse set of solutions with an HBR first and subsequently find the best skill for a situation with Bayesian Optimisation.

\section{Preliminaries and Motivation}

\subsection{Reset-free Trial and Error}
The Reset-free Trial and Error algorithm (RTE) ~\cite{Chatzilygeroudis2018Reset-freeRecovery} is a powerful method to enable a robot to adapt to changing environments or scenarios in real time. First, a repertoire of solutions is created by MAP-Elites in simulation where each solution $\theta$ is stored with respect to the observed behavioural descriptor $\mathbf{bd}_\theta$. 
After creating the repertoire in simulation, it is used as a prior for the mean of a set of Gaussian processes. The main loop of RTE consists in running a Monte Carlo Tree Search (MCTS)~\cite{remi2006mcts} algorithm to plan the next best action $\mathbf{bd}_{t+1}$ together with the Gaussian processes given the actual state of the robot $s_t$. Since the simulation environment is always imperfect, Gaussian processes map the behavioural descriptors $\mathbf{bd}_\theta$ to observed behaviours $\mathbf{bd}_{observed}$ in the new environment. Once the action from the repertoire is executed, the observed behavioural descriptors are used to update the Gaussian process before we do the next round of planning.
After this initialisation, the main loop of RTE runs until a stop criterion is met, e.g. reaching the goal state. 

RTE has been successfully used to enable a hexapod robot and a velocity-controlled differential drive robot to reach its final destination even with one or more damages. The algorithm takes advantage of the diversity of skills created by the MAP-Elites algorithms and shows to be very effective at finding solutions in the repertoire of diverse solutions that are suitable for different situations. In contrast to ITE, RTE has more diverse skills but only one way to execute them. This behaviour implies that in an unexpected situation the algorithm needs to have a viable solution in the repertoire for each skill which is not always the case. For this reason, an effective algorithm relies on a good diversity of skills with a redundancy of ways to execute these skills. On one hand, ITE promotes the diversity of executing a single skill and on the other hand RTE pushes for a diversity of skills without a diverse way of executing them. Our method aims to optimize for both properties to make the adaption of robots more effective.




\subsection{Hierarchical Behavioural Repertoires}\label{hbr_back}

To increase the number of ways to achieve different skills, \hbr{} extends the Hierarchical Behavioural Repertoire (HBR) algorithm \cite{Cully2018} to learn locomotion skills for robots. HBR algorithms show to be effective at finding complex skills that can be used across different robots (e.g. drawing digits with a robot-arm and a humanoid robot \cite{Cully2018}) without retraining all the layers. 
Structurally, HBRs consist of layers of MAP-Elites repertoires which are chained together to create diverse skills. A bottom layer could   consist of a controller that makes a robotic arm move to any reachable point in a defined space. Following this architecture, we can train a second layer that draws a line from one point to another by reusing solutions from the first layer (i.e. reach two points with the arm). These new skills can be used by a third layer to create arcs with three different lines. Finally, the robotic arm can learn how to draw digits by drawing different arcs with the skills from the third layer.
All the layers are trained sequentially (i.e. one after the other) from the bottom up since each repertoire of behaviours (i.e. layer) is using previous behavioural repertoires to create more complex skills. In the special case of only using one layer, the HBR algorithm reduces to a classical QD algorithm with a single repertoire. HBR algorithms are able to create complex solutions (e.g. draw digits with a robotic arm) by building on top of previously built repertoires. \hbr{} uses the HBR algorithm to optimise its skills in different sub spaces (i.e. different behavioural repertoires) of the full solution space for the complex skills it needs to adapt during the deployment.

\textit{\textbf{Behavioural Descriptor and Genotype.}}
Each layer $k$ in \hbr{}, contains solutions $\theta_k$ with a behavioural descriptor $\mathbf{bd_k}$. The solutions are defined by their genotype $\mathbf{g}_k$ where $\mathbf{g} \in G$.
Since multiple layers are stacked on top of each other, each repertoire needs to be connected to the others. In HBRs, this is done by using a genotype $\mathbf{g}_k$ definition that maps to the behavioural space of the other layers.

\textit{\textbf{Stacking.}}
The solutions $\theta_1$ in the lowest layer correspond to the parameters we use to directly control a robot. For the other layers $k$, the solutions $\theta_k$ in layer $k$ use the behavioural descriptor space of the lower layers as the action space. In practice, this is implemented as a succession of behavioural descriptor coordinates $\mathbf{bd}_{k-1}$ which is defined in the genotype $\mathbf{g}_k$ of a solution. This definition allows to produce a successive execution of the corresponding controllers in the lower layers $k-1$. 
In other terms, the middle layer is a mapping $\phi():B_2 \rightarrow B_1$ from the behavioural descriptor space of the middle layer to an output space, which corresponds to the behavioural descriptor space of the lower layer.

Hierarchical repertoires are not limited by the number of layers and can become increasingly complex by simply stacking more layers of repertoires. 

\section{Methods}
\label{sec:methods}

\begin{table}
    \centering
    \begin{tabular}{p{2cm}|c c c}
                                 & Top Layer & Middle Layer & Lower Layer  \\
        \hline
        Discretisation      & 100x100 Grid & 0.05 & 0.01 \\
        Mutation Rate       & 0.14   & 0.11 & 0.17  \\
        Generations         & 20000 & 30000 & 5001 \\
        Genotype Size       & 9 & 18 & 6
    \end{tabular}
    \caption{QD Parameters used for \hbr{} with a population of 200 individuals. The mutation is polynomial with $\eta_m$ and $\eta_c$ both at 10.0. The discretisation of the behavioural space is defined by a grid (Top Layer) or an $l$ value. }
    \label{tab:qd_params}
\end{table}
\subsection{Hierarchical Architecture}
Our Hierarchical Trial and Error (\hbr{}) algorithm leverages the hierarchical repertoires to improve the diversity of skills, decompose complex solution spaces and make the adaption to different damage scenarios more effective. 
Overall, the \hbr{} algorithm uses a three-layered HBR (see Fig.~\ref{fig:hbr_architecture}) where (i) the bottom layer defines the movement for a single leg for 1 second, (ii) the middle layer makes the robot walk for 1 second by selecting 6 controllers for the legs and (iii) the top layer makes the robot walk for 3 seconds by chaining 3 controllers from the middle layer.

\textit{\textbf{The \hbr{} Architecture.}}\label{sec:arch}
With the \hbr{} algorithm, our goal is to find behaviours that control the robot in various directions while doing it in many different ways. 
To this end, the \textbf{bottom} layer is directly controlling the legs of the hexapod. A solution in that repertoire is an open-loop controller for one single leg. 
Each leg of the hexapod has three motors, where the motor attached to the body uses a periodic function $\varphi_1(t,a,p,d)$ at each time-step $t$ with parameters $a$ for the amplitude, $p$ for the phase shift and $d$ for duty-cycle. The two other motors use the same periodic function $\varphi_2(t,a,p,d)$ such that the legs are always perpendicular to the bottom.

We use $a_1,p_1,d_1$ to control the first motor and $a_2,p_2,d_2$ to control the second and third motors. This configuration means that the \textit{genotype} consists of 6 parameters namely, $g_1=\{a_1,p_1,d_1,a_2,p_2,d_2\}$. 
The \textit{behavioural descriptor} $\textbf{bd}_{l}$ for the bottom layer is measured at the end of a simulation of 1 second for each leg. The descriptor is defined as the height $h$ of the move, the swing distance $d_{swing}$ and the duty cycle $d$ of the second and third motor. The duty cycle $d$ in this case is both a parameter and a \textit{behavioural descriptor}.
To encourage minimal energy consumption, the \textit{fitness function} is set as the sum of the commands that are sent to the motors throughout the simulation.

For the \textbf{middle} layer, the \textit{genotype} consists of selecting 6 leg behaviours from the bottom layer $g_2=\{\textbf{bd}_{1},\textbf{bd}_{2},\dots,\textbf{bd}_{6}\}$ where $\textbf{bd}_{l}$ is a behavioural descriptor vector for solutions of the bottom layer from a specific leg where $l \in \{1,2,3,4,5,6\}$.
The \textit{behavioural descriptor} has 3 dimensions and consists of the $x$, $y$ and $yaw$ displacement of the robot after one second.
The \textit{fitness function} measures the angular distance between the robot's final orientation and an ideal circular trajectory (similar to existing works~\cite{Cully2013}). The resulting repertoire consists of controllers that enable the robot to walk in any direction for one second.

Lastly, in the \textbf{top} layer, the \textit{genotype} consists of 3 behavioural descriptors from the middle layer, $g_3=\{\textbf{bd}_1,\dots,\textbf{bd}_3\}$. 
The \textit{behavioural descriptor} of this top layer is the final $x$, $y$ position (i.e. $x$, $y$ displacement) of the robot which is measured after 3 seconds of simulation. 
Our \textit{fitness function} is similar to the middle layer namely the angular distance between a perfect trajectory and the robot's orientation.

\textit{\textbf{Primary and Secondary Behavioural Descriptors.}}\label{cond_behaviours}
To find a larger diversity of solutions in QD algorithms, it is possible to make the behavioural space $B$ larger by adding additional dimensions to it. However, this can quickly lead to an exponential growth of the repertoire size which will make it i) challenging to cover the behavioural space due to the decrease of evolutionary selection pressure that happens with large collections \cite{7959075} and ii) difficult to store it in memory even on modern computers. In hierarchical repertoires, if the size of the behavioural space $B$ on lower levels increases, we can achieve more combinations for the high level skills (and thus increase the diversity). The difference between both approaches, is that lower layers in hierarchical repertoires can store fewer and simpler solutions to cover an equivalent behavioural space than a traditional single-layered, or "flat", repertoire. This means that the growth in size is significantly lower than it is for complex solutions that are trained in a traditional repertoire.
To distribute different information flows to different subsystems in the hierarchy (i.e. information factorisation \cite{Merel2019}) in HBRs we can increase the size of our behavioural space $B$ in the middle layer of the HBR by increasing $B$ with additional \textbf{secondary} behavioural descriptor $\mathbf{bd}_{\textrm{secondary}}$ dimensions. The dimensions of the behavioural descriptor which were originally described in the previous section are called \textbf{primary} dimensions and they remain unchanged. The additional dimensions of our behavioural space are \textbf{secondary} which means that higher level layers do not have to define these as part of their genotype when selecting solutions from that lower repertoire. The secondary dimensions can be used to execute solutions with the same \textbf{primary} behavioural descriptors but in different ways. 
The training of the upper layers remains unchanged by directly selecting the solution in the lower layer with the nearest primary behavioural descriptor from the genotype value, without considering the secondary descriptor. Doing so, reduces the search space for the top level and helps the optimisation process to find diverse skills.

In our implementation of \hbr{} for the hexapod, we extend our middle layer with 6 additional dimensions for the secondary descriptors. In addition to the 3 existing primary dimensions $\textbf{bd}_{\textrm{primary}}$, we now have the secondary behavioural descriptor values $\textbf{bd}_{\textrm{secondary}}$ which measure the ground contact for each leg (1 if it is in contact with the ground for more than 30\% of the time, otherwise 0). Since we have 6 legs, this gives us 6 additional secondary dimensions with 64 different possibilities (see Fig.~\ref{fig:hbr_architecture}). The top layer will keep the same genotype as it focuses only on \textbf{primary} behavioural descriptors during the training. The \textbf{secondary} descriptors will be used by \hbr{} during the adaptation phase to select the most appropriate way to execute each skill given the condition of the robot. 

\begin{figure}[h]
\centering
\includegraphics[width=0.45\textwidth]{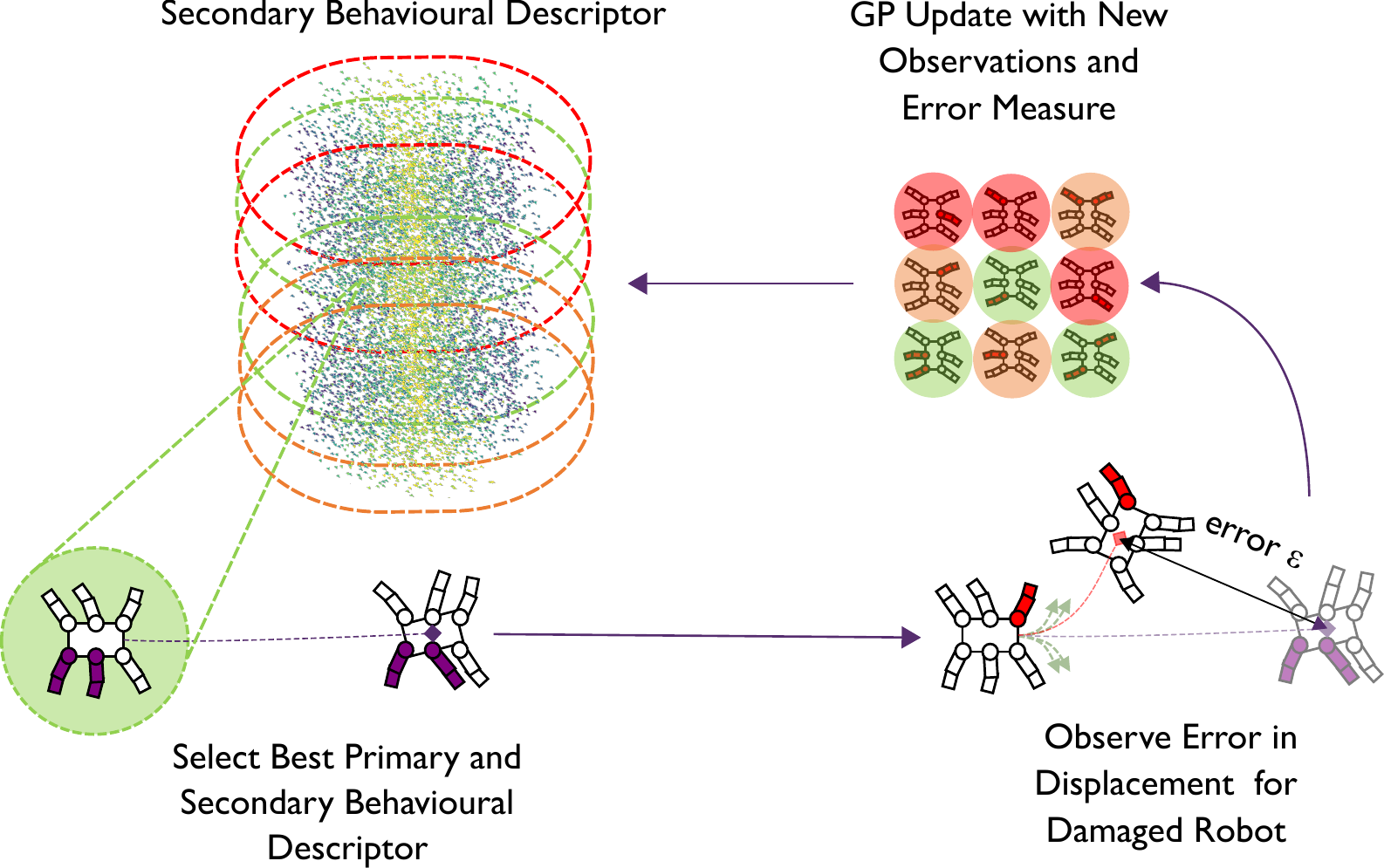}
\centering
\caption{Trial and Error Selection of Secondary Behavioural Descriptors. First, we select a secondary descriptor (purple leg) and a primary descriptor ($x$ and $y$ displacement) and execute it. Second, we observe the error of behaviour on the damaged robot (red leg) and lastly we update the Gaussian processes to find unaffected secondary descriptors.}
\label{fig:gp-ucb}
\end{figure}

\subsection{Online Hierarchical Skill Adaptation}
We are interested in leveraging the diversity that we have incorporated into the hierarchical architecture to adapt the behaviour of the robot to different situations in the context of a maze navigation task. The adaption is done by selecting skills that are able to recover their behaviour for a specific situation (e.g. damages). 

\textit{\textbf{Primary Behaviour Selection via RTE.}}
HTE combines the adaptation and planning capabilities of RTE with the skill expressivity of the HBR. To this end, RTE uses the HBR in \hbr{} like a traditional single-layered repertoire by exclusively interacting with the top layer. RTE chooses which skill should be executed from the repertoire to come closer to the next goal by using MCTS and GPs. Since \hbr{} can execute each high-level skill with different secondary behavioural descriptors, we do not only need to choose which high-level skill to execute with RTE but we also need to select a secondary behavioural descriptor to adapt to the situation. 
 
 \textit{\textbf{Secondary Behaviour Selection via Trial and Error.}}\label{sec:trial}
To find the optimal secondary descriptor and adapt to an unforeseen situation, \hbr{} minimises the following error measure to find the secondary behavioural descriptors whose primary behaviours are the least affected by the situation:
 
\begin{equation}
\mathbf{\epsilon} = \exp{(-k\frac{|\mathbf{bd}^\prime-\mathbf{bd}_\theta|}{2|\mathbf{bd}_\theta|-c})}
\label{eq:error}
\end{equation}
 
where $\mathbf{bd}^\prime$ is the observed primary behavioural descriptor and $\mathbf{bd}_\theta$ the desired one. Both are the behavioural descriptor vectors with the $x$, $y$ coordinates of the robot's centre of mass and the desired yaw rotation  ($\mathbf{bd} = \{x, y, yaw\}$) (see Sec.~\ref{sec:arch}), and $k,c$ are  positive hyper-parameters to scale and shift the data respectively. In our setting we used $c=0.5$ and $k=4$. This error measure is inspired from the error measure introduced by APROL~\cite{Kaushik2020AdaptiveRobotics}.

To minimise the error $\epsilon$ (Eq.~\ref{eq:error}) we use Bayesian Optimisation with Gaussian processes (GPs). GPs are a family of stochastic processes that are particularly attractive for this kind of regression problems because of their data-efficiency and uncertainty quantification. The GPs will learn to predict $\epsilon$ at every step for all combinations of a high-level skill and a secondary behaviour. Finally, to select the optimal secondary descriptor, we will use the upper-confidence bound (UCB) acquisition function~\cite{Srinivas2009GaussianDesign}. The UCB method will pick the secondary descriptors that minimise the error $\epsilon$ given the current situation of the robot. This process is done during the deployment of the robot in the new situation without any resets (see Fig. \ref{fig:gp-ucb}). At each iteration of the trial and error iteration in \hbr{}, the GPs are updated with the new observed behaviours to refine their predictions of the least affected secondary descriptors.
In \hbr{}, we have a finite set of possible secondary behaviours and thus we can iterate over them to find the best high-level skill with RTE and the best secondary behavioural descriptor for a situation with UCB.


\begin{figure}[h]
\centering
\includegraphics[width=0.35\textwidth]{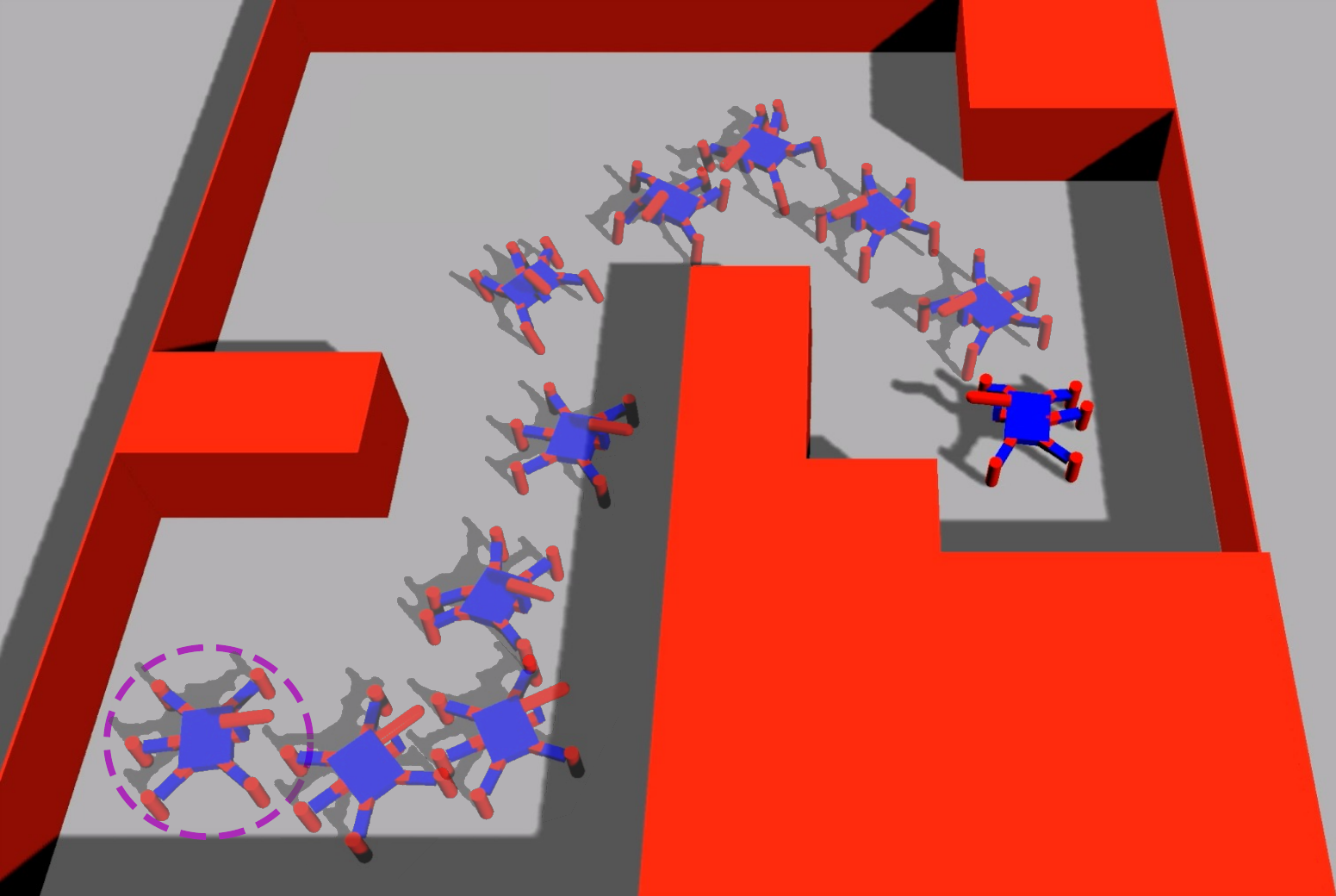}
\centering
\caption{Maze Navigation Task. The damaged (left middle leg) hexapod robot needs to reach the purple circle in a least amount of steps.}
\label{fig:environment}
\end{figure}

\section{Experimental Evaluations}
 \subsection{Experimental Setup}
 We evaluate the benefits of \hbr{} (with parameters in Table \ref{tab:qd_params}) in a maze solving task (see Fig. \ref{fig:environment}) with the hexapod being damaged in several ways. We create 7 different tasks, where each tasks consists of solving the maze but each time the robot has a different leg that is damaged (i.e. a leg is blocked in the air). Since we use a hexapod we damage each leg individually and the 7th task consists of damaging both middle legs at the same time. 
 
 \begin{figure}[h!]%
\centering
\includegraphics[width=0.47\textwidth]{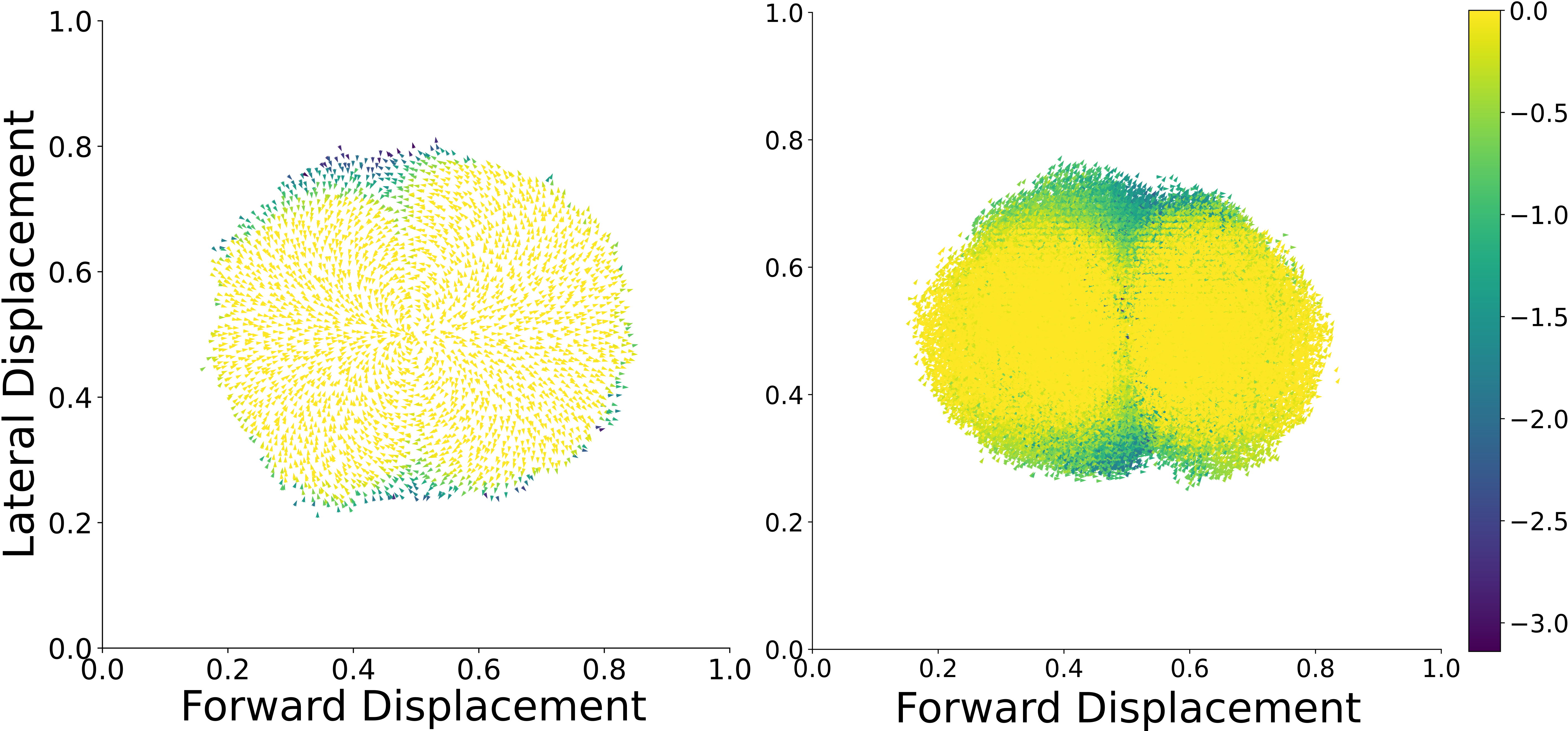}
\caption{The left repertoire belongs to the \hbr{} method and the right one to the 8-dimensional flat variant. Both show the $x$ and $y$ location after 3 seconds of movements for the robot on a flat plane. The right repertoire has a higher density since we collapse 8 dimensions onto 2 to compare them. The legend shows the fitness of the solutions.}
\label{fig:archives}
\end{figure}

 In the following experiments we consider three baselines, namely RTE with a 2-dimensional repertoire (2D-RTE) \cite{Chatzilygeroudis2018Reset-freeRecovery}, RTE with a 8-dimensional repertoire (8D-RTE) and APROL \cite{Kaushik2020AdaptiveRobotics}. The "flat" repertoires we use for RTE should have the same solution space (and same behavioural space) as the HBR used in \hbr{}. The open-loop controllers for the flat repertoires are defined by a genotype with 36 parameters and they are executed for 3 seconds. For this purpose, the flat 8-dimensional repertoire version will have a behavioural descriptor space of $2+6$ dimensions where each controller is executed for 3 seconds. The first 2 dimensions correspond to the final $x$, $y$ displacement of the robot, while the other 6 dimensions represent a binary leg contact similarly to the middle layer of \hbr{}. For the 2-dimensional version, we use the final $x$, $y$ displacement as behavioural descriptor which is the same we use for the top layer of \hbr{} (see Fig. \ref{fig:archives}).
 
 For the baselines, we use the code from  the original authors of RTE, and we re-implemented APROL in C++ with the SferesV2 framework \cite{Mouret2010}. 
 For the planning method within APROL, we use MCTS and A* similarly to RTE which differs from the original implementation where only A* was used. For APROL we created 21 different flat 2-dimensional repertoires where each repertoire has been trained with a single or double leg damage on a flat terrain. For the flat repertoires used in RTE and APROL, we use the same hyper-parameters than in the original RTE paper.
 For the simulations, we use the framework \emph{robot\_dart} \footnote{\url{https://github.com/resibots/robot_dart}} which is an accessible wrapper that allows to create robot tasks in the DART \cite{Lee2018} simulator. The implementations for HTE and the experiments are made available at \url{https://github.com/adaptive-intelligent-robotics/HTE}.
 
 To compare the different repertoires, we created 4 repertoires for each variant and replicated the evaluations 30 times for each damage.


\begin{figure}[h]
\includegraphics[width=0.34\textwidth]{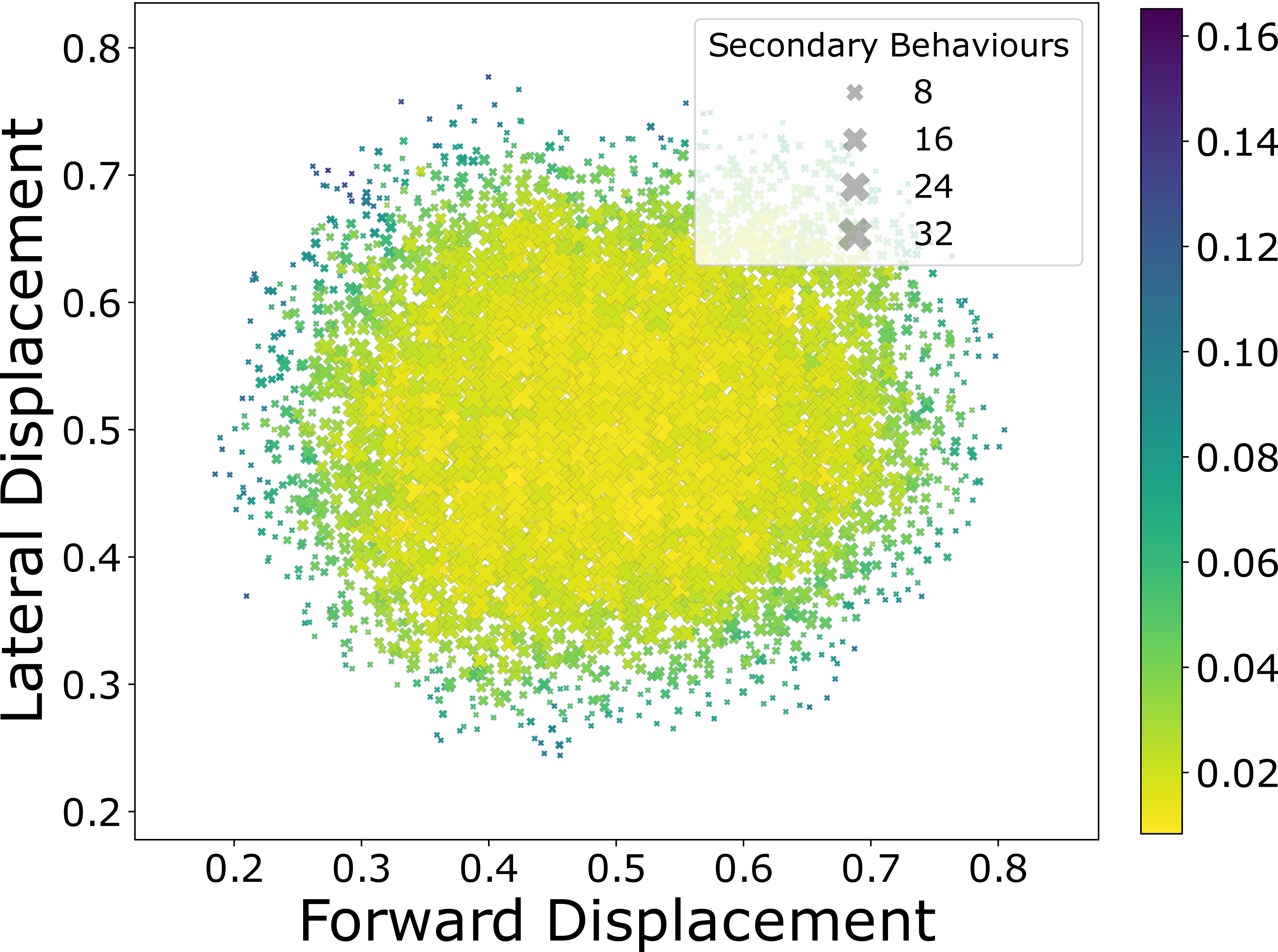}
\caption{Final $x$, $y$ position the hexapod robot can reach while using different secondary behavioural descriptors. The size of the crosses represents the number of secondary behaviours from the middle layer that we can execute without changing the behavioural descriptor of the solution in the top-layer repertoire. The colour represents the median error between the desired $x$, $y$ position and the robot's final position.}
\label{fig:conditional_behaviours}
\end{figure}

\subsection{Can the hierarchical repertoire be more diverse than a regular flat repertoire?}

We want to compare the diversity of solutions produced by HBR, compared to the one produced by the 8D repertoire.

 \begin{table}
\centering
 \begin{tabular}{c c c c} 
 \toprule
    Algorithm & Repertoire Size & Effective Size & Mean Fitness  \\ [0.5ex] 
 \midrule
 \hbr{} & $2980.8 \pm 131.0$ & $2980.8 \pm 131.0 $&$-0.01 \pm 0.02 $\\
 
Flat-8D & $72528.0 \pm 2809.4$ &$2531.5 \pm 111.8$ &$-0.21 \pm 0.02 $\\
 \bottomrule
\end{tabular}
\caption{Results for each architecture (at $4e6$ evaluations). The numbers  of solutions in the repertoire differ a lot since the flat repertoire has 8 dimensions and the HBR version only has 2. The \textit{Effective Size} is the number of cells we fill when projected on the first two dimensions and the \textit{Mean Fitness} is the mean fitness of these individuals.}
\label{table:hbr_results}
\end{table}
We present the results in Table~\ref{table:hbr_results}, where we observe that the flat version is able to find a high number of solutions. This result is expected since we have $640 000$ cells that can be filled. However, when projecting the content of the repertoire on the two first dimensions, namely the $x$, $y$ displacement, we can observe from Fig.~\ref{fig:archives} that \hbr{} covers a larger space than the flat 8D repertoire. 
We report the coverage of this projection in Table \ref{table:hbr_results} as the \textit{Effective Size} which represents the number of cells that we have filled.
In these comparisons, \hbr{} has a better "effective" coverage and a better average fitness than the flat 8D repertoire. This demonstrates that \hbr{} can generate a better repertoire for the x, y displacements which are the skills we are interested in for the navigation tasks.

\begin{figure}[h]
\includegraphics[width=0.35\textwidth]{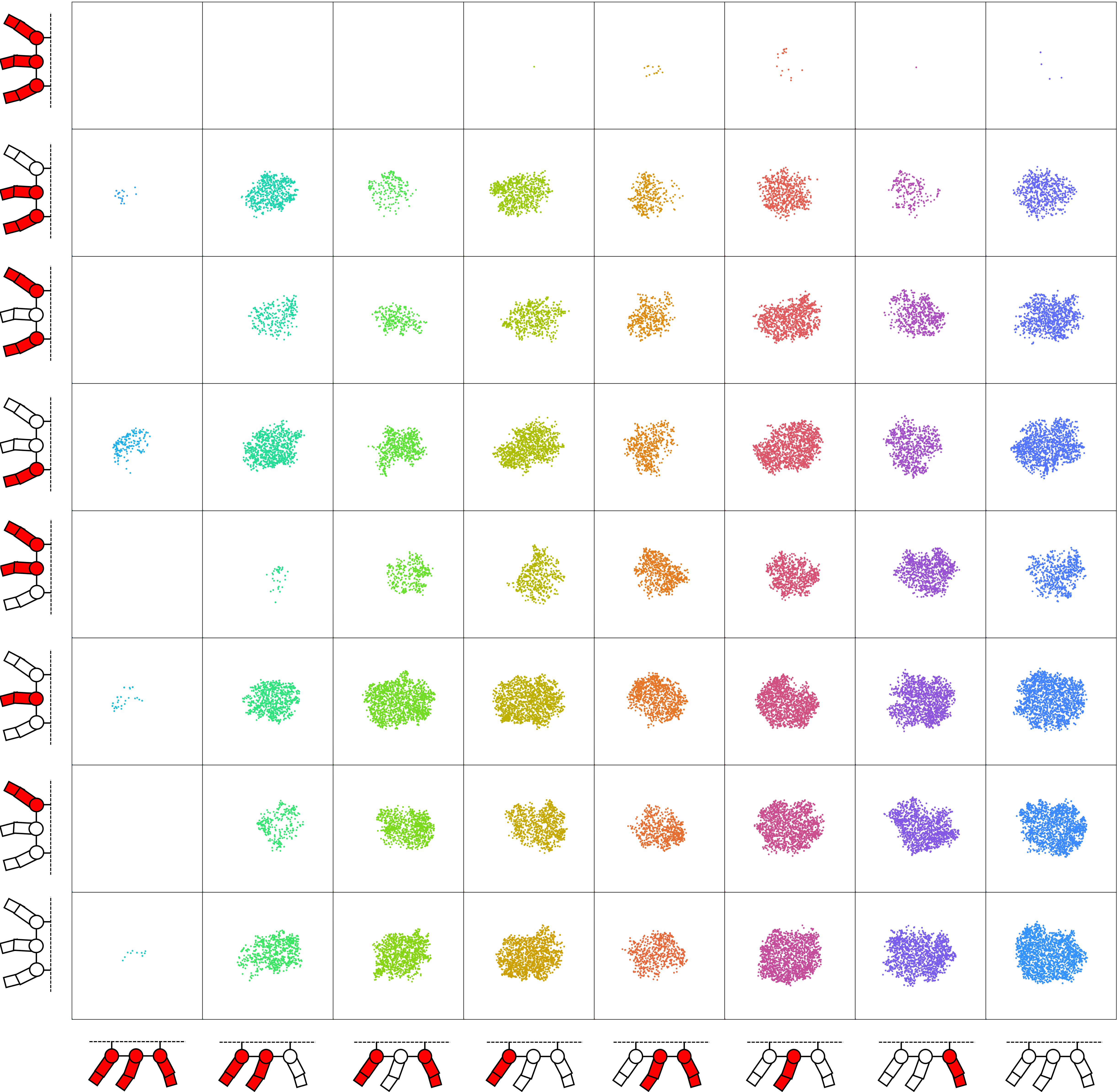}
\caption{Each subplot shows the final $x$, $y$ positions we can reach with the hexapod while using different leg duties (i.e. more or less than 30\% contact with the ground). On this plot we have one subplot per desired secondary behavioural descriptor where each leg is red if it is not used for that subplot.}
\label{fig:conditional_behaviours_grid}
\end{figure}

\subsection{Can we modulate high-level skills with secondary behavioural descriptors?}
To test whether we can modulate high-level skills, we re-evaluate all solutions from the top-layer of \hbr{} with the objective to reach the same behaviour descriptor (i.e. $x$ and $y$ position) while using different secondary behavioural descriptors in the middle-layer. 
For testing purposes, we use the same secondary behavioural descriptor for the full duration of the skill (i.e. the secondary behaviour will be the identical for the 3 steps) which leaves us with 64 possible choices for each high-level skill. Some of these secondary behaviours are expected to be impossible to achieve for the robot (e.g. to not use any of its legs). After re-evaluating the solutions, we count how many times the robot (i) exhibited the wanted secondary behaviour and (ii) whether the robot was able to reproduce its behaviour descriptor by reaching its original final $x$, $y$ position.

In Fig.~\ref{fig:conditional_behaviours}, we plot how many times the robot reaches the desired $x$, $y$ location after 3 seconds while executing different secondary behavioural descriptors. The colour of the legend tells us how close to a desired $x$, $y$ location we get in the median case for executed movements that match the desired secondary behavioural descriptor. The size of the markers is proportional to the numbers of high-level skills that can be executed with secondary descriptors. On Fig.~\ref{fig:conditional_behaviours}, we have a big region of low error in the centre with a large numbers of valid secondary descriptors. Therefore, the robot is able to move to various $x$, $y$ positions while using differently its legs in a multiple ways. 
We can observe that the error grows and the number of secondary behaviours decreases near the edges of the repertoire. This inverse correlation can be expected as it is likely more challenging for the robot to find a large number of different ways to walk fast in different directions.

In Fig.~\ref{fig:conditional_behaviours_grid} each subplot corresponds to the final $x$, $y$ positions reachable with a specific secondary behavioural descriptor (64 in total). This result shows which secondary descriptors are actually not feasible (e.g. the first column or first row of the subplots). 
On both figures, we can observe that \hbr{} is able to use secondary behavioural descriptors to modulate its high-level skills toward moving to the same point while exhibiting a different behaviour even though we have not expanded the behavioural space on the top layer. These results show that \hbr{} successfully reuses the diversity within the hierarchical structure.

\begin{figure}[h]%
\centering
\includegraphics[width=0.38\textwidth]{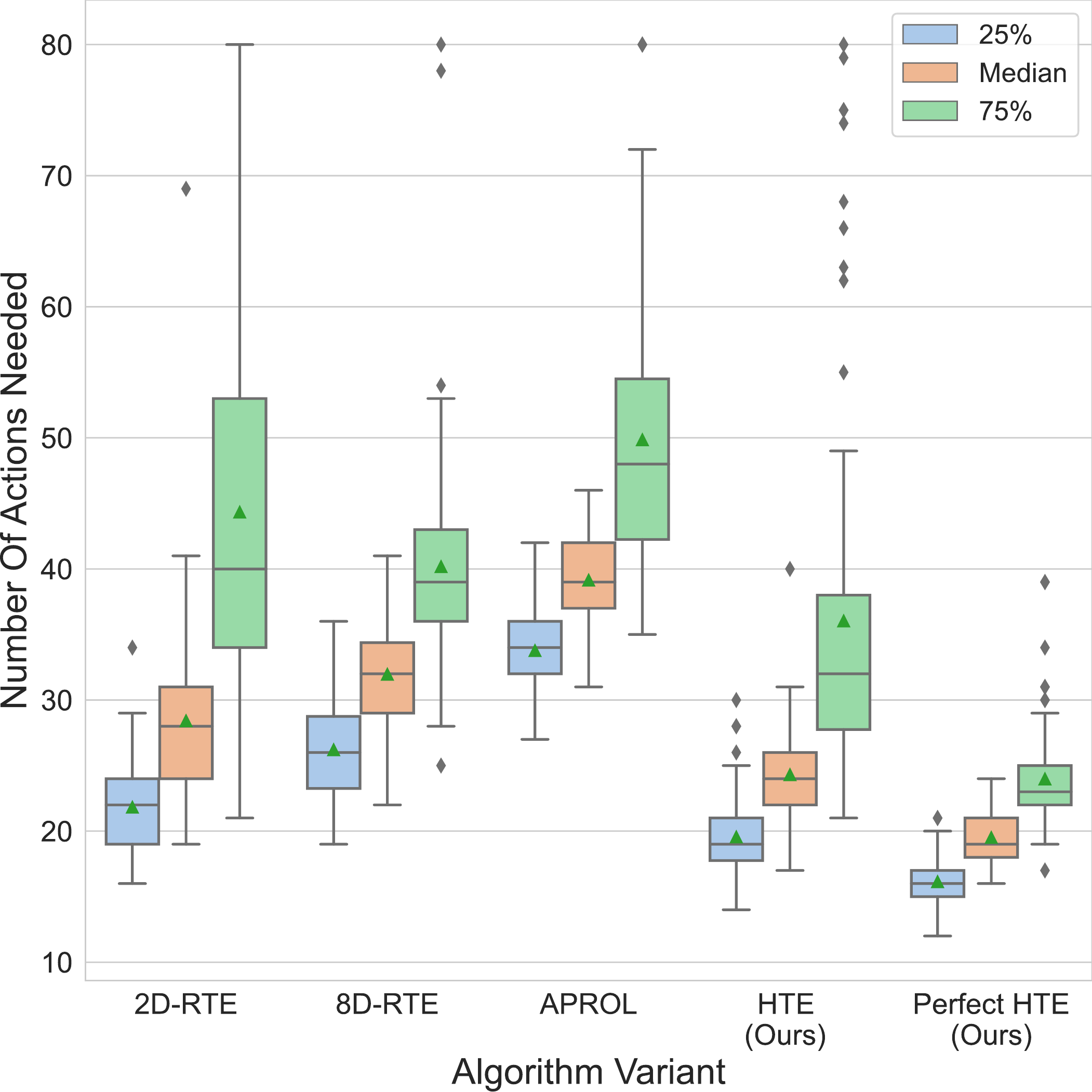}
\caption{Box-Plot for the different variants of algorithms for which we report the 25, 50 and 75 percentile values for each replication.}
\label{fig:actions_per_repertoire}
\end{figure}

\subsection{What are the benefits of \hbr{} on the robot's adaptation capabilities?}
Finally, we evaluate the benefits of \hbr{} and the additional behavioural diversity provided by the secondary behavioural descriptors on the robot's adaptation capabilities. More specifically, we test the algorithm in a maze navigation task with a damaged hexapod (see Fig.~\ref{fig:environment}).

We run \hbr{} in two different scenarios. First, an upper-baseline scenario where the damage of the robot is known and we can directly select a preferred secondary behaviour (called Perfect \hbr{}). Second, a scenario where we let the robot adapt automatically by choosing the optimal secondary behaviours for the given situation with Bayesian Optimisation (see Sec. \ref{sec:trial}).
We are interested in knowing which version is the most adaptive across different scenarios. Thus, we report the distribution of the median numbers of actions that a single variant replications needs across the 7 tasks. Since we did 30 replications per repertoire we obtained $7*30=210$ median values for our report per variant. Moreover, we are interested in how the variants perform in the best-case and worst cases which is why we also report the $25th$ and $75th$ percentiles in addition to the medians.

From the results in Fig.~\ref{fig:actions_per_repertoire}, we can see that the best performing algorithm is \hbr{} with prior knowledge. The manual selection of the secondary descriptor based on the prior knowledge of the damage helps the robot to adapt. However, it is expected that this prior knowledge and expertise might not always be available. When this information is not available, \hbr{} still performs better than all the other variants, even if it has to infer from experience the optimal secondary behaviours for recovery.

\begin{figure}[h]%
\centering
\includegraphics[width=0.4\textwidth]{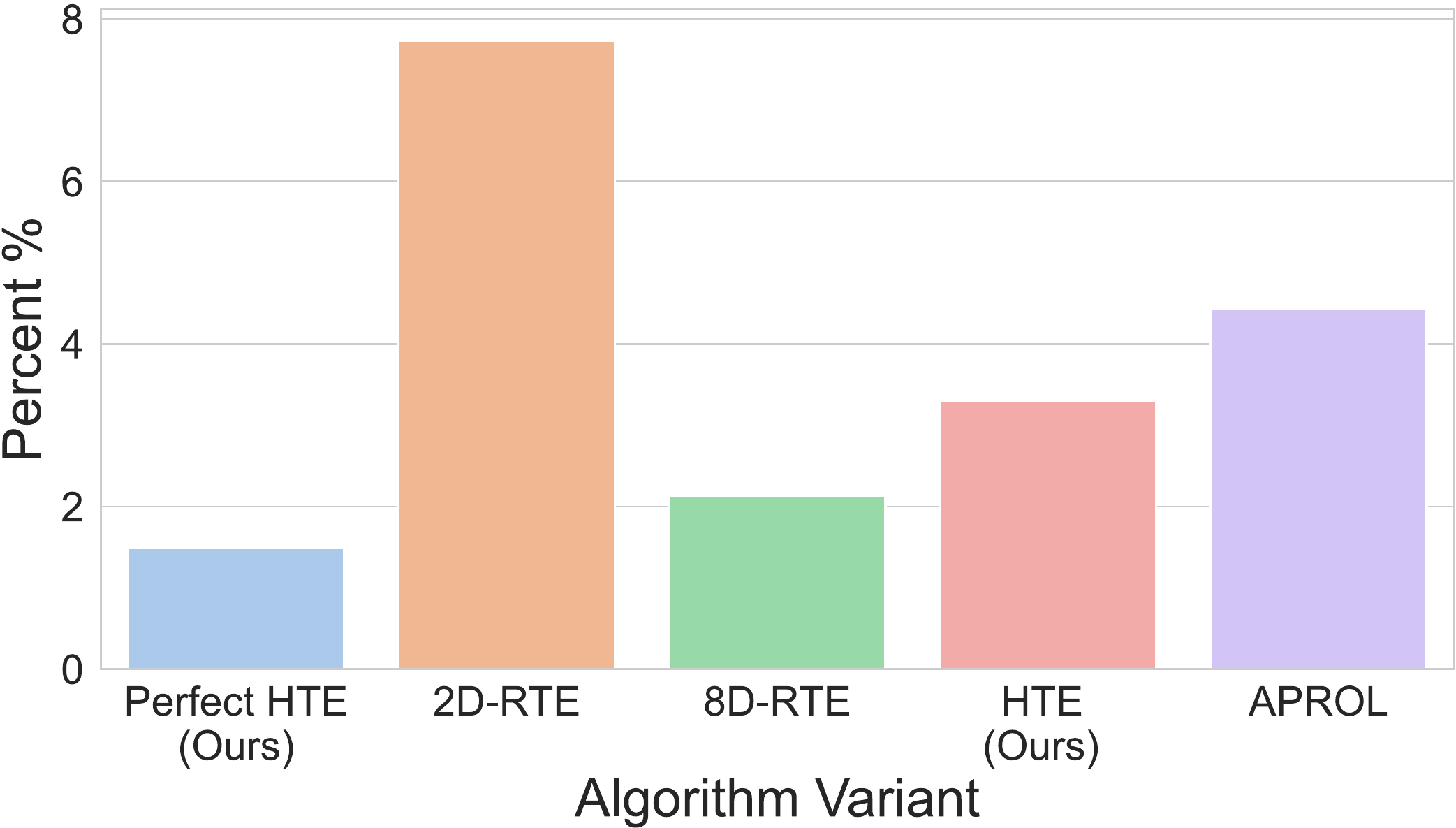}
\caption{Percentage of the experiments that failed to finish in under 80 actions for each variant.}
\label{fig:found_percentage}
\end{figure}

In median, we can observe across the replications that \hbr{} needs 24 actions, while 2D-RTE needs 28, 8D-RTE 32 and finally APROL 39 actions. This result shows a $14.2\%$ improvement for \hbr{} over the best baseline, namely 2D-RTE for the median cases.
We can see from the results in Fig.~\ref{fig:actions_per_repertoire} that RTE with a 2D repertoire performs very strongly for some cases. For example, for the 25th percentile we only need $22$ actions in median to solve the 7 tasks for the 2D-RTE variant and $19$ for the HTE variant. APROL has the worst performance in the experiments. This might be due to a reduced number of available repertoires (we use 21 repertoires instead of 57 repertoires in the original paper as we do not study the effect of friction). Another difference is that we use MCTS as a planner instead of A* from the original APROL study. For APROL, we see that even in the best scenarios we need $34$ actions and for 8D-RTE we need $26$.

\hbr{} performs well even in the worst case scenarios (i.e. the $75th$ percentiles) where it takes $32$ actions in median to solve the tasks in comparison to $40$ for the 2D-RTE, $39$ for 8D-RTE and $48$ for APROL. In these worst case scenarios, we show a $20\%$ improvement over the best baseline (2D-RTE). However, \hbr{} still fails to match the performance of the Perfect \hbr{} variant which only needs $16$ actions for the 25th percentile, $19$ for the median and $24$ for the $75th$ percentile of replications. Perfect \hbr{} shows that there is still room for the design of a better secondary behaviour selection in future work. 
The results on the downstream task for each algorithm are statistically different according to a Wilcoxon–Mann–Whitney test and a p-value < 1e-6 with a Bonferroni correction.

In a few cases, the robot may flip on its back and get stuck or stall in a corner of the maze. For these reasons, we limit the number of possible actions the robot can take to solve the navigation task to 80 (i.e. a failure). While this hard-coded value could bias the average of the results, the reported box-plots for the median values are less influenced by outliers. For the $25th$ and $75th$ percentile values, we do not use an interpolation for the same reasons, but we take the closest data point that corresponds to the $nth$-percentile value. 
In Fig.~\ref{fig:found_percentage}, we show the failures across replications of each variant for the task. \hbr{} has $57\%$ less failures in the maze experiments when compared to the best performing baseline in the maze (2D-RTE). The 8D-RTE and APROL versions are more stable by having less failures, but they need significantly more steps to solve the task. \hbr{} is able to both solve the maze quickly while being more stable with regard to total failures.

The results show that the diversity created by \hbr{} helps in scenarios where a robot requires to find many diverse ways of executing a behaviour to improve its damage recovery capabilities.

\section{Conclusions and Future Work}
In this paper, we introduced the Hierarchical Trial and Error algorithm, which allows robots to leverage the diversity of hierarchical repertoires for damage recovery. \hbr{} splits the complexity of the behavioural space across different layers and introduces primary and secondary behavioural descriptors. This results in a repertoire that can store more diversity than other baselines without making the training process more complex. Finally, we can effectively use this diversity in an online fashion to recover from damages while solving a maze navigation task. The results from the experiments show that the damage recovery by \hbr{} needs $14\%$ less actions than the best baseline in the median cases and, more importantly, \hbr{} is more robust against complete failures. 

One of the main limitations of this work, is the definition of behavioural spaces on lower layers which heavily impact the upper layers by restricting the search space. If the lower layers don't have enough diverse solutions, \hbr{} will not find optimal solutions for the upper layers and thus \hbr{} requires more expertise to define different genotypes, behavioural descriptors and fitness functions.

Interesting directions for future work could involve an improved skill selection mechanism to chose the best secondary descriptors. The results show that the perfect \hbr{} has the best adaptability capabilities but the skills need to be chosen correctly.
\begin{acks}
This work was supported by the Engineering and Physical Sciences Research Council (EPSRC) grant EP/V006673/1 project REcoVER. We want to also thank the members of Adaptive and Intelligent Robotics Lab (AIRL) for their valuable inputs.
\end{acks}

\bibliographystyle{ACM-Reference-Format}
\bibliography{references}

\end{document}